# Design and Fabrication of a Microinjector to Conduct Columnar Injection of Stem Cells for Treating Parkinson’s Patients

Erfun Bayat Sarmadi*
*School of Mechanical Engineering*
*College of Engineering*
*University of Tehran*
Tehran, Iran
bayat.sarmadi.er@ut.ac.ir

Sayed Mahdi Same*
*School of Mechanical Engineering*
*College of Engineering*
*University of Tehran*
Tehran, Iran
mahdi.same@ut.ac.ir

Mahdi Abolghasemi-Dehaqani
*College of Engineering*
*University of Tehran*
Tehran, Iran
mhdi.abolghasemi@gmail.com

Arash Bahrami
*School of Mechanical Engineering*
*College of Engineering*
*University of Tehran*
Tehran, Iran
arash.bahrami@ut.ac.ir

Mohammad-Reza Abolghasemi-Dehaqani
*School of Electrical & Computer Engineering*
*College of Engineering*
*University of Tehran*
Tehran, Iran
dehaqani@ut.ac.ir

Aghil Yousefi-Koma
*School of Mechanical Engineering*
*College of Engineering*
*University of Tehran*
Tehran, Iran
aykoma@ut.ac.ir

**Abstract— Considering the rising prevalence of neurological diseases like Parkinson's, novel treatment methods have advanced significantly in recent years. One promising approach is transplanting stem cells into targeted brain regions, for which developing precise equipment is essential. A critical method for implanting these cells is columnar injection into brain tissue. The present research focuses on designing and fabricating a microinjector device for the controlled columnar injection of stem cells into the brains of Parkinson’s patients. Inspired by existing microdrive models, the newly developed device stands out for using two motors as actuators, two encoders, and a PID controller. This configuration compensates for operational errors and environmental noise, ensuring precise injections at the required rate. The PID controller is designed using system identification, where equations governing the power transmission systems are derived and implemented. To verify the device's performance, the injection rate for a standard procedure (25 microliters at 1 microliter per minute) is measured using encoders and validated with a digital caliper. The instantaneous injection rate is calculated using the encoder and caliper data. While the average of these data is the same, the standard deviation of the caliper data is significantly lower than that of the encoder data. This indicates that the device is functioning as expected and that much of the noise in the encoder data originates from the sensor itself.**

## I. INTRODUCTION

Parkinson's disease cases have increased by 1% annually from 1990 to 2019, but in 2022, North American cases grew by 50% compared to the previous year [1, 2], signaling a rapid acceleration. This has driven researchers to explore new treatments [3], particularly stem cell-based therapies [4, 5]. These therapies require specialized equipment for a specific injection [6, 7], called columnar injection, a technique that causes less tissue damage than bolus injection [8], as shown in Figure 1.a. A comparison between the mechanism of bolus and columnar injection is illustrated in Figure 1.b. In the bolus injection method, the syringe needle and body remain stationary while the plunger is pressed downward to inject into a specific point. In the columnar injection method, however, the syringe body and plunger move upward. The body moves faster than the plunger, allowing the stem cells to be gradually released in a columnar pattern and the needle to be drawn out of the tissue.

To conduct the columnar injection, Schweitzer and colleagues added a mechanism to the FHC microdrive device [9], which is used for neurosurgical procedures [10]. This system is displayed in Figure 2. The main issue with this method is the system's open-loop nature and the inability to adjust the injection rate due to having only one actuator. Consequently, changing the injection rate requires regulating the mechanism. Also, performance errors caused by friction,

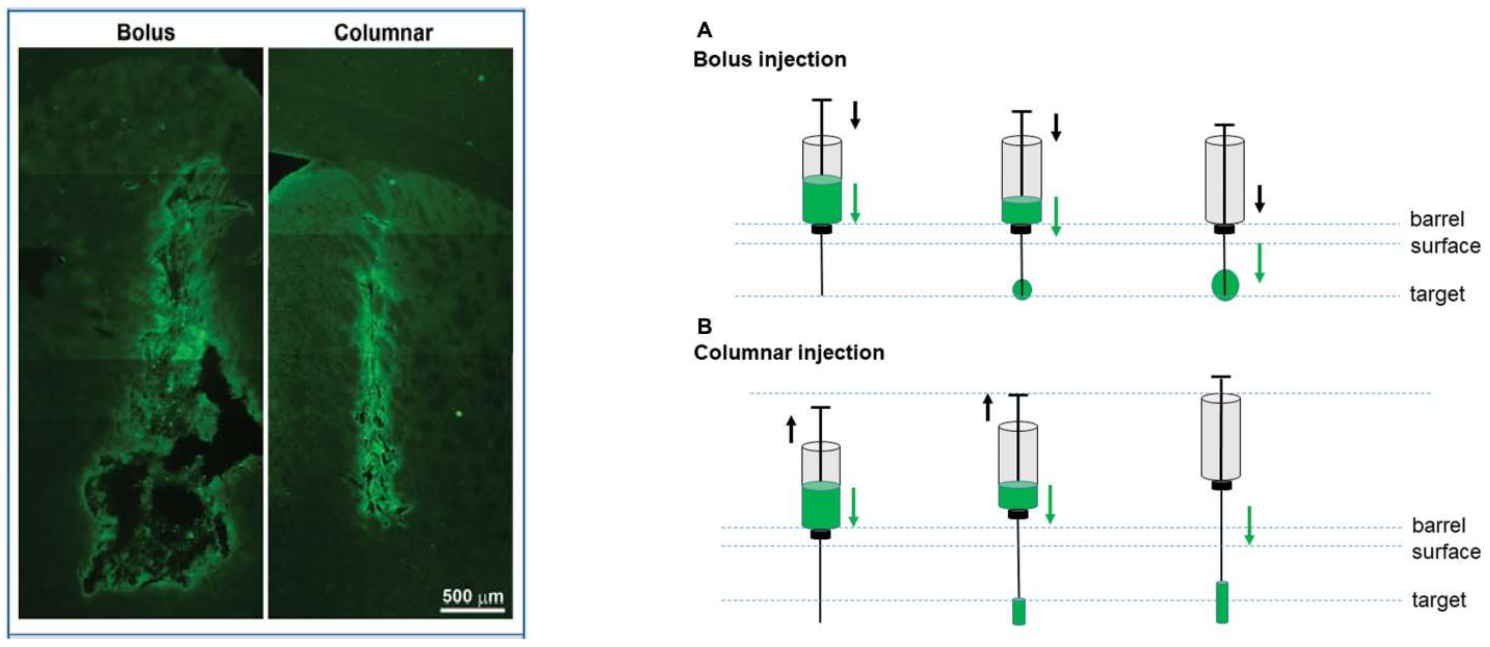

a. Damaged Caused to Tissue

b. Mechanism

Figure 1. The comparison between Bolus and Columnar Injection [9]

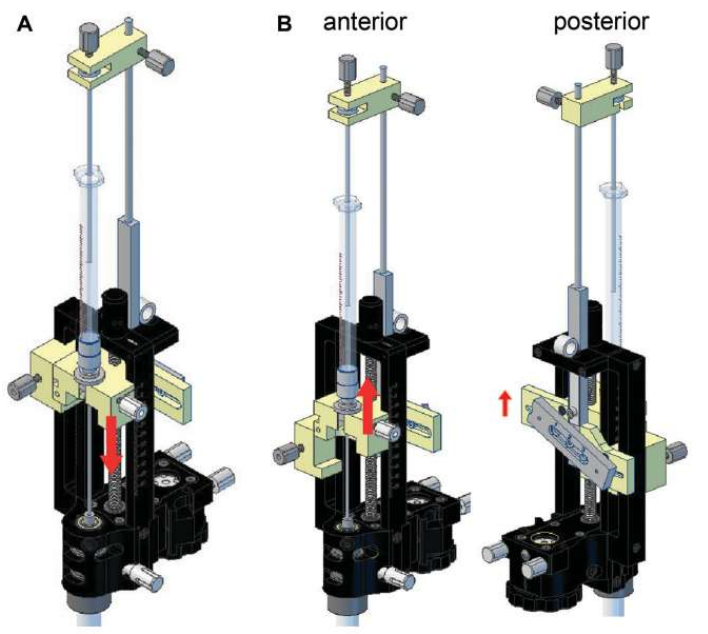

Figure 2. The Device Proposed By Schweitzer et al. to Conduct Columnar Injection [9]

*Authors Contributed Equally.

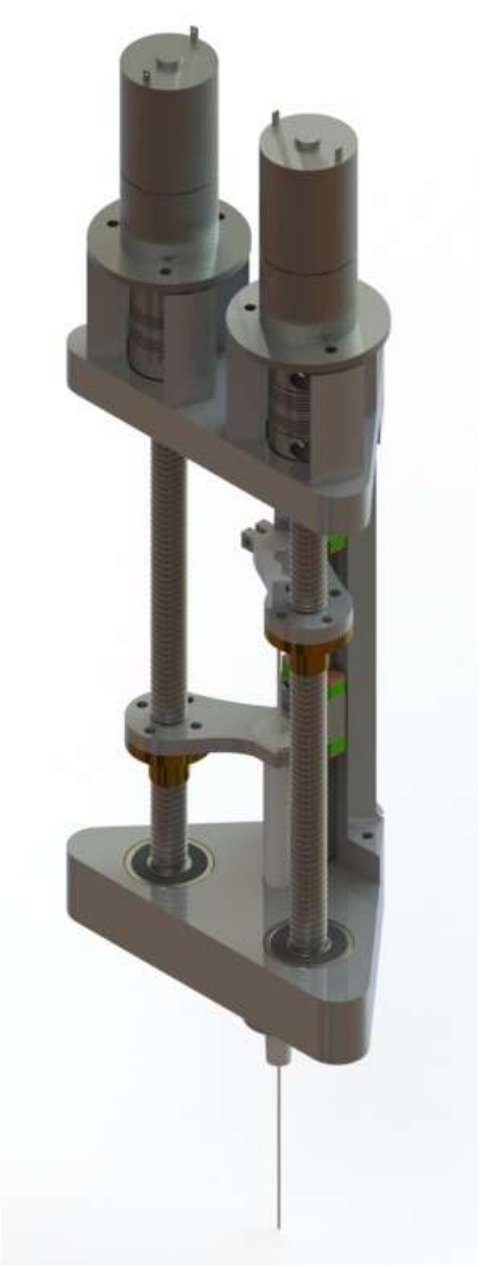

a. Isometric View

| ITEM NO. | PART NAME | QTY. |
|---|---|---|
| 1 | DC motor | 2 |
| 2 | Coupling | 2 |
| 3 | Upper Frame | 1 |
| 4 | Bearing | 4 |
| 5 | Plunger Connection Interface | 1 |
| 6 | Barrel Connection Interface | 1 |
| 7 | Lead Screw | 2 |
| 8 | Lead Screw Nut | 2 |
| 9 | Syringe | 1 |
| 10 | Colunmar Body | 1 |
| 11 | Wagon | 2 |
| 12 | Rail | 1 |
| 13 | Lower Frame | 1 |
| 14 | Encoder | 2 |

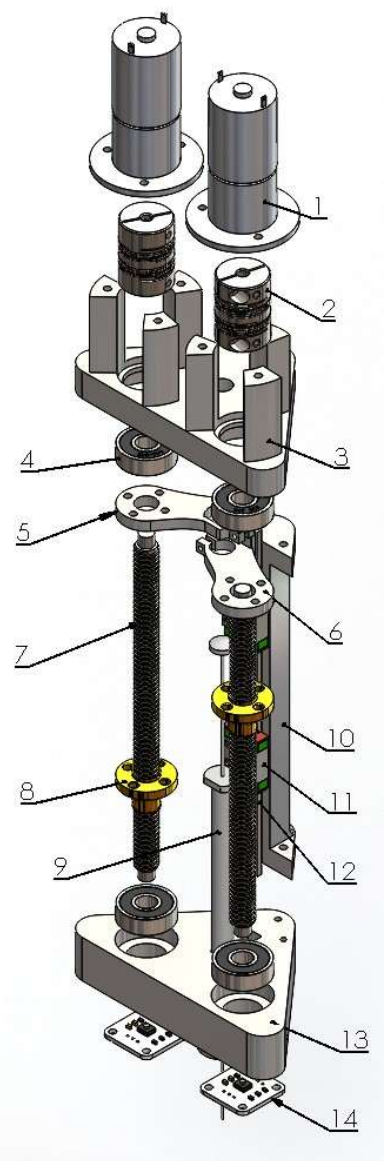

b. Exploded View

Figure 3. 3D View of the Device

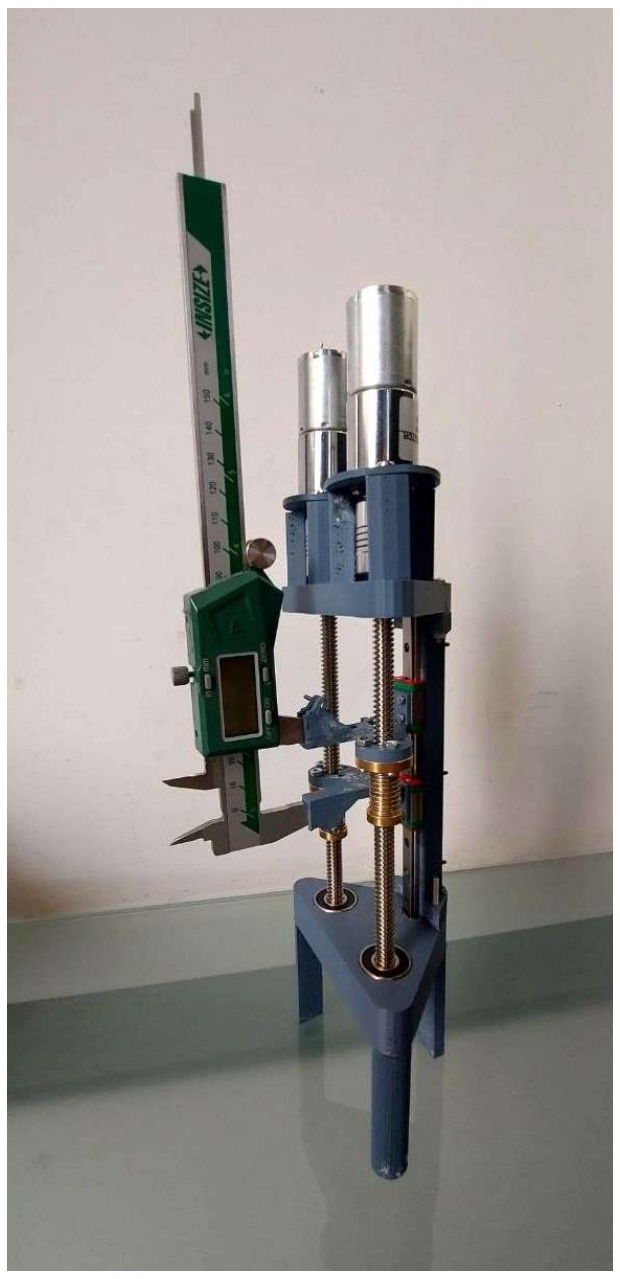

Figure 4. 3D Printed Prototype

voltage fluctuations, and environmental noise can affect injection precision. The objective of this research is to design and build a microinjector that addresses these issues using sensors, a controller, and two DC motors as actuators.

## II. DESIGN & FABRICATION

The design and construction of the microinjector includes four layers:

### A. *Mechanisms Layer*

The mechanisms layer consists of the design and fabrication of the microinjector device. This device includes two power transmission systems, each comprising a DC motor, ball bearings, a lead screw, and a nut. Each lead screw nut is connected to a wagon via a connecting interface to prevent the nut from rotating. In this way, the linear motion of the nuts is emerged. While one nut is attached to the syringe plunger, the other one is connected to the syringe body. This setup allows separate control of the syringe body and plunger movement by each motor. In the columnar injection method, if both the plunger and barrel of the syringe move upward simultaneously, the needle is retracted from the patient's brain as the injection proceeds, which is favorable for neurosurgeons. Additionally, the limitations related to the range in which the motors can be controlled can be either eliminated or smoothed out.

In addition, a part mounted below the device's base plate serves as the interface between the microinjector and the stereotactic device, which supports the microdrive equipment used in neurosurgery. The isometric view and the exploded view of the microinjector are shown in Figure 3.

In a typical injection, 25 microliters of stem cells are injected into the brain at a rate of 1 microliter per minute, taking 25 minutes. Given that the syringe's piston diameter is 1.03 mm and the lead screw pitch is 2 mm, the angular velocity difference of the motors is determined in (1).

$$v_{rel} = \frac{|\dot{V}|}{A} = \frac{1}{\frac{\pi \times 1.03^2}{4}} = 1.200\,\frac{\text{mm}}{\text{min}}$$

$$\omega = \frac{v_{rel}}{p} = \frac{1.200}{2} = 0.6\ \text{rpm} \qquad (1)$$

Where $\omega$ is the angular velocity difference of the motors, and $v_{rel}$ is the relative speed of the syringe body movement compared to the plunger. Also $p$ denotes the lead screw pitch, Moreover, $\dot{V}$ represents the rate of decrease in the injected volume or injection rate, and $A$ is the cross-sectional area of the syringe piston. Equation (1) shows that the angular speed difference between the two motors must be 0.6 rpm to achieve the desired injection rate. Later, we discuss that the motors used in the device can be controlled at velocities ranging from 0.5 to 1.5 rpm. Therefore, the injection process is feasible.

Since the encoders produce noisy results, a digital caliper is used as a distance measurement sensor for data validation. As the description suggests, it measures the distance between the parts connected to the syringe plunger and the body at all times. In the next versions of the device, this caliper will be replaced with a displacement sensor.

This device is fabricated using Fused Deposition Modeling (FDM) with 3D printers to reduce costs and fabrication time. The material used is polylactic acid (PLA). The printed components are then assembled to complete the microinjector, as shown in Figure 4.

### B. *Electronics Layer*

The electronics layer consists of the design of the electronic circuit of the device. The electronic components used in the device are presented in Table 1.

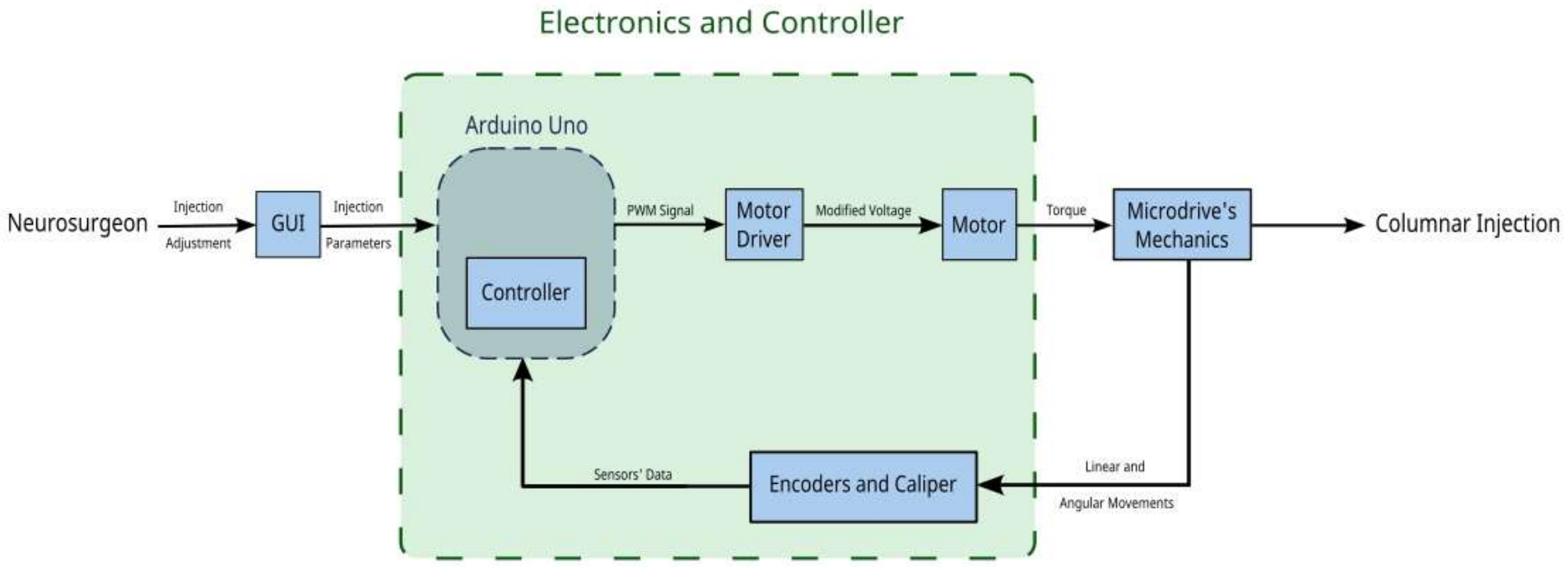

Figure 5. Circuit Schematics

Table 1. Electronic Parts List

| Electronic Part | Model |
|---|---|
| Development Kit | Arduino Uno |
| DC Motor | GA25RP |
| Motor Driver | L298N |
| Magnetic Encoder | AS5600 |
| Distance Sensor | Insize 1108-150 |

As shown in Figure 5, the encoders and the caliper are connected to the Arduino kit to measure the rotation angle of screws and the distance between nuts respectively. The DC motor drivers are also connected to the Arduino and provide enough voltage and current to drive DC motors. Due to high computational cost and processing requirements, algorithms and data processing are performed using MATLAB.

### *C. Controller Layer*

The primary goal of this layer is precise control of the injection speed while ensuring patient safety. The control system manages the injection process accurately and in a coordinated manner using data from sensors and a MATLAB application. This section examines the various components of the control system, the feedback system, and how each part functions.

In this project, digital PID control is preferred over continuous PID control because the device's processing and control unit is a laptop, which is well-suited for running digital algorithms. Additionally, digital PID control allows for the simple and dynamic adjustment of control coefficients, offering more flexibility in tuning and optimizing the system's performance.

In reality, due to the fluid's viscosity, the movement of the plunger can affect the movement of the syringe body and vice versa. However, modeling this phenomenon requires a deep understanding of the stem cell's mechanical properties including density and viscosity. Since the synthesized amount of stem cells produced by the research institutes in charge is negligible and hard to access, there is no straight way of measuring the mentioned properties. Therefore, for simplicity, each motor, along with the magnetic encoder and lead screw, is considered an independent dynamical system. For each of them, in order to estimate its dynamic model, an input voltage in a stepwise manner ranging from 0 to 5 volts is applied. The input is shown in Figure 6.

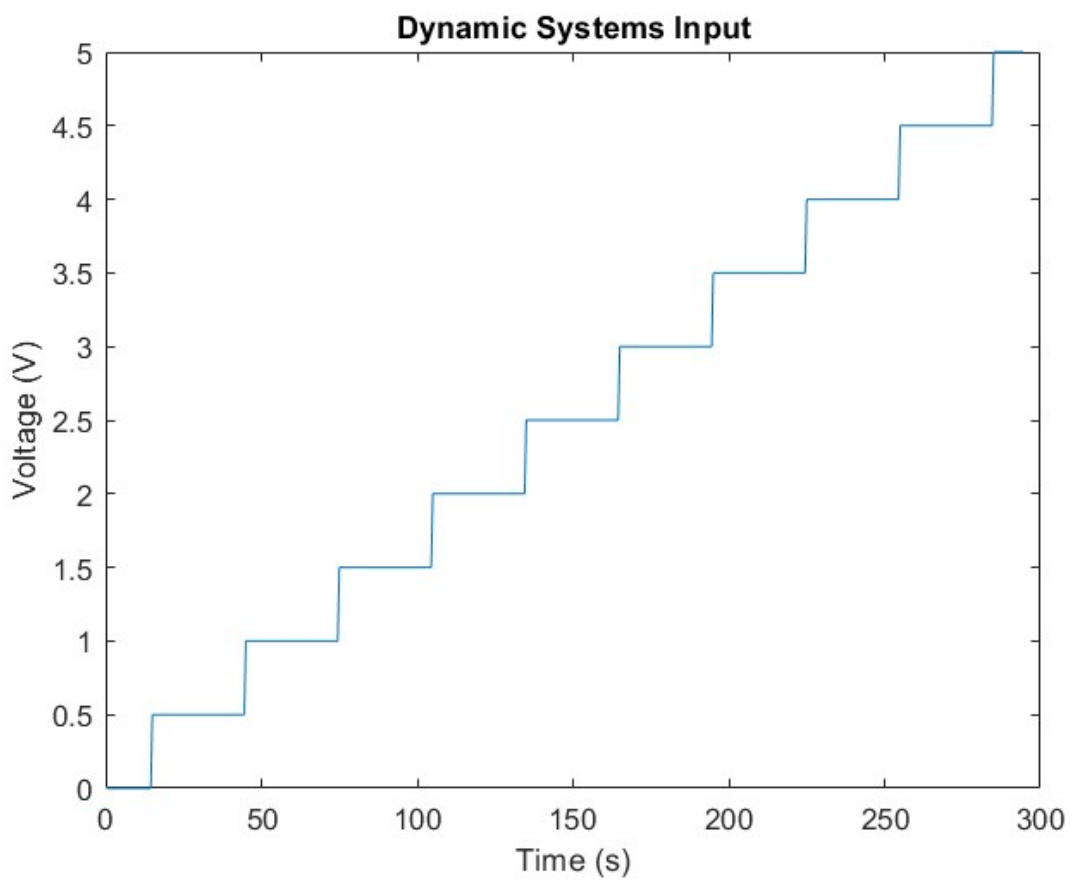

Figure 6. System Identification Input

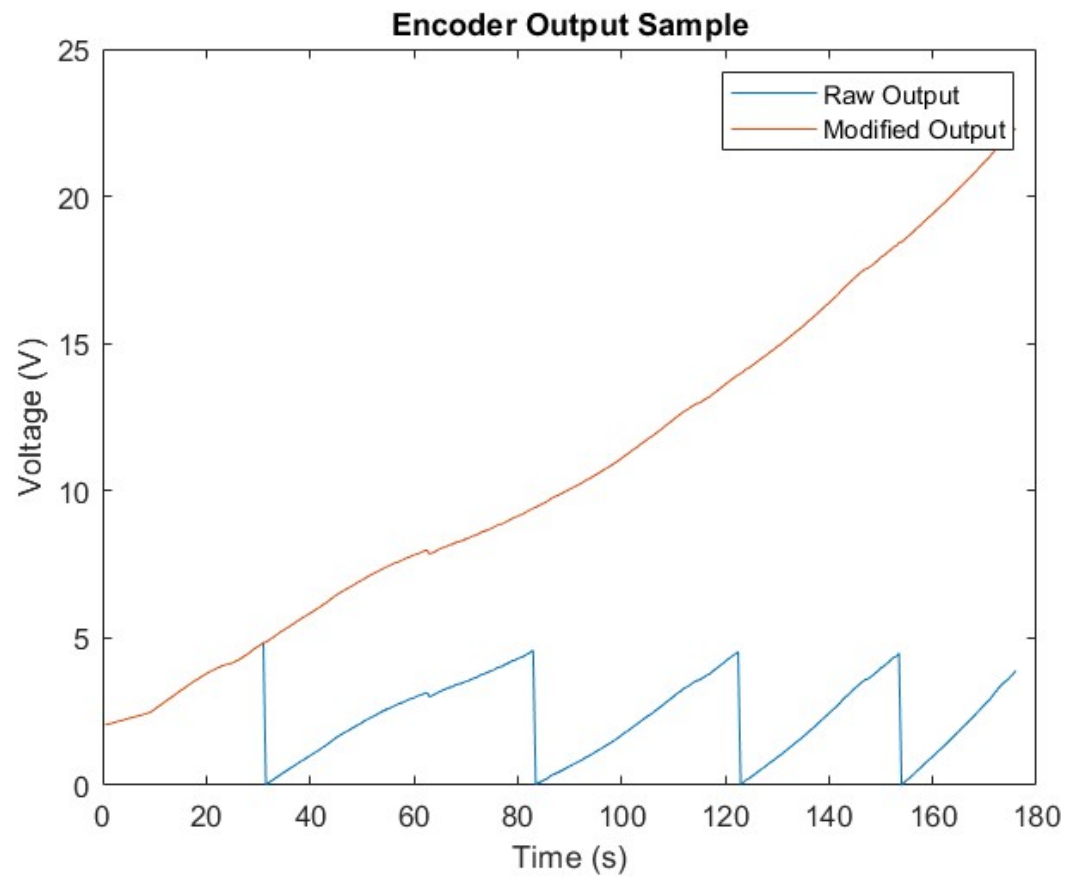

Figure 7. A Comparison Between Raw and Modified Output

Table 2. System Identification Results

| System | Transfer Function | Fit to Estimate Data | MSE |
|---|---|---|---|
| Body DC Motor | $\dfrac{0.02147\, z^{-1} - 0.02139\, z^{-2}}{1 - 2.391\, z^{-1} + 1.789\, z^{-2} - 0.3981\, z^{-3}}$ | 94.82% | 0.1542 |
| Plunger DC Motor | $\dfrac{-0.004654\, z^{-1} + 0.02205\, z^{-2} - 0.01737\, z^{-3}}{1 - 1.904\, z^{-1} - 0.08905\, z^{-2} + 1.893\, z^{-3} - 0.8996\, z^{-4}}$ | 94.78% | 0.2514 |

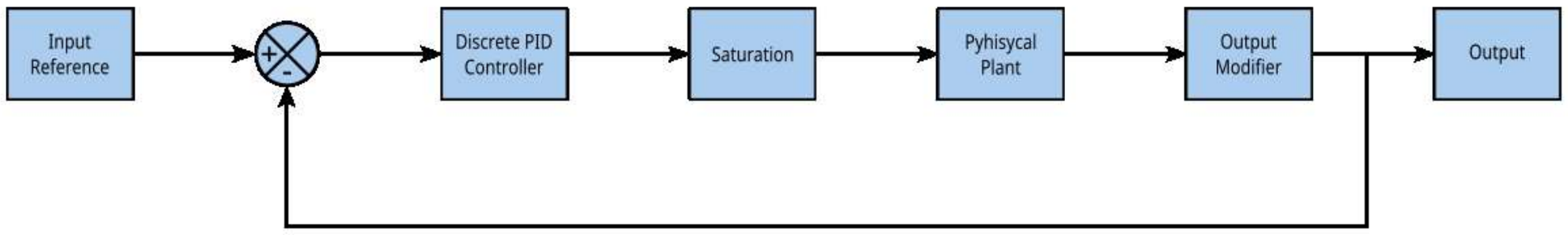

Figure 8. Control Loop Structure

The output from the encoders is absolute and is recorded and stored. This output varies as an analog signal from 0 to 5 volts. After a full rotation, the signal suddenly drops from 5 volts to 0 or suddenly rises from 0 to 5 volts. Considering this and the natural measurement noise that contaminates the signal, the numerical derivation of the output signal is unreliable and cannot be used as feedback. Therefore, there is no choice but to use the position control method, which requires the real-time modification of the encoders' output. This process is done using an algorithm written in MATLAB. A sample of the processed signal along with the raw output from the angle sensors is displayed in Figure 7. As shown in this figure, the modified output of the encoders is the accumulated voltage of the raw inputs. The slope of the modified output is the angular velocity of the DC motors. Considering this, by using the position control method, the problem can be categorized as a reference tracking problem since feeding a ramp input to the system maintains its constant angular velocity.

After storing the stepwise input and modified output of each open-loop system, they are fed to MATLAB System Identification to estimate the transfer function of the systems and proceed to model-based controller design. The estimated models are tabulated in Table 2.

As the first step of the model-based controller design, the closed-loop control structure is designed and shown in Figure 8. As appears in this figure, the difference between a ramp input and the modified output of each encoder is fed to a PID controller which determines the voltage input to its corresponding DC motor. A saturation block is employed to ensure that the plant input voltage remains within the 0 to 5 volts range. The output modifier block is also applied to modify the output of the physical system. Based on the mentioned structure, a PID controller was designed for each dynamic system using the MATLAB PID Tuner App. While the controllers worked well in simulations, they failed to perform appropriately in real systems due to the Arduino's limited voltage supply. This left no choice but to fine-tune them for physical implementation. Eventually, an identical controller was employed for both motors. The primary controller used in simulations and the physically implemented controller are shown in Table 3. Using this controller, the angular velocity of each DC motor can be controlled in a range of between 0.5 and 1.5 rpm.

It is worth mentioning that the encoders are embedded in the control loop, while the digital caliper data are not used in the loop and are solely used to validate the data obtained from the encoders.

### *D. GUI Layer*

In the microinjector device, the graphical user interface (GUI) plays a crucial role in communication between the users and the device. The medical team, including physicians and nurses, requires a simple and intuitive interface to operate the device easily and with high confidence. The designed GUI is displayed in Figure 9. Some of the benefits of having a GUI include:

- Reducing the risks associated with human errors
- Allowing quick adjustments and control of the device with minimal effort
- Facilitating the learning and familiarization process for the medical team

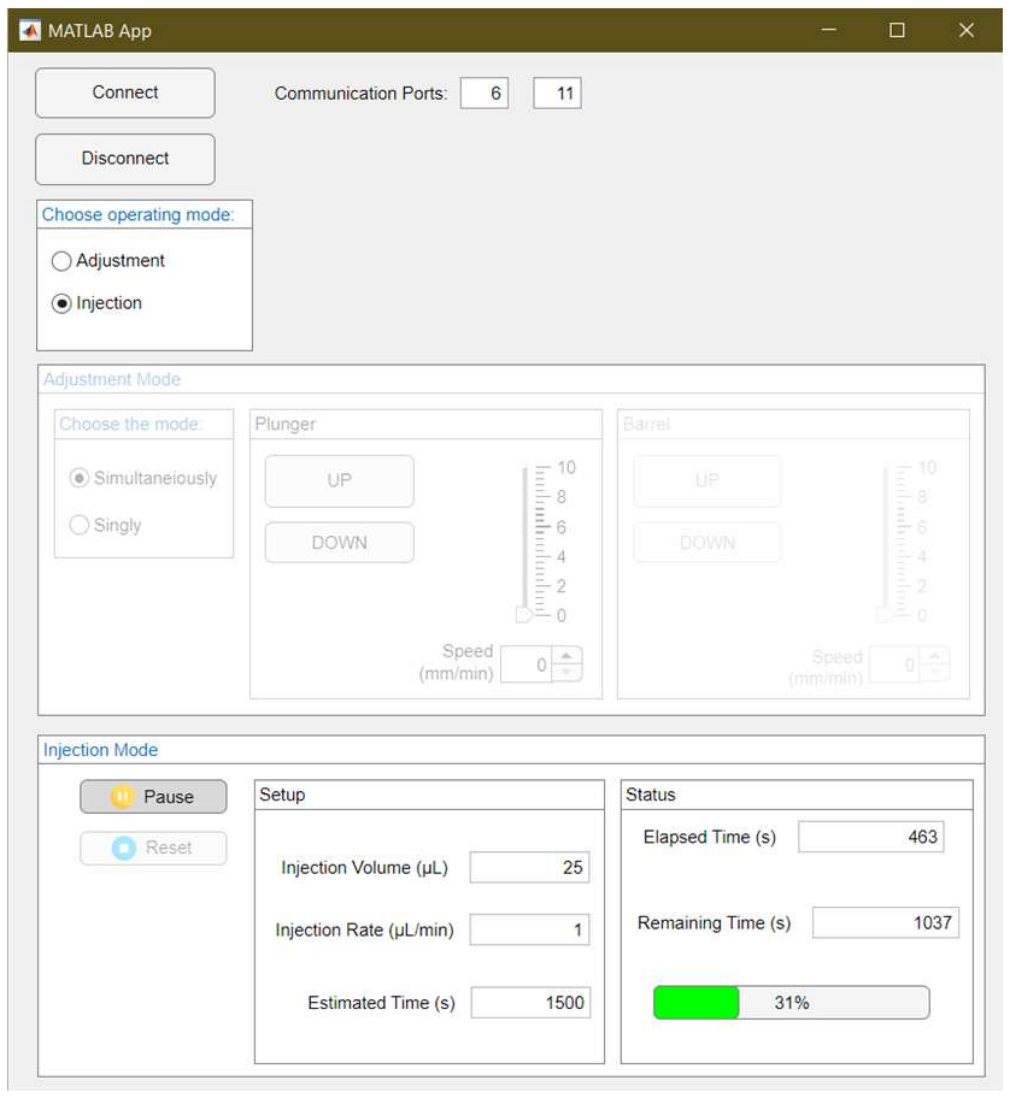

Figure 9. Graphical User Interface During the Injection

Table 3. Digital PID Controllers' Coefficients

| Controller | $k_p$ | $k_i$ | $k_d$ | N | $T_s$ |
|---|---|---|---|---|---|
| Body DC Motor | 2.0194 | 0.079197 | 5.0445 | 0.32467 | 0.5 |
| Plunger DC Motor | 0.67902 | 0.0057207 | -0.067005 | 2.8663 | 0.5 |
| Physically Implemented | 5.21580 | 0.033700 | - | - | 0.5 |

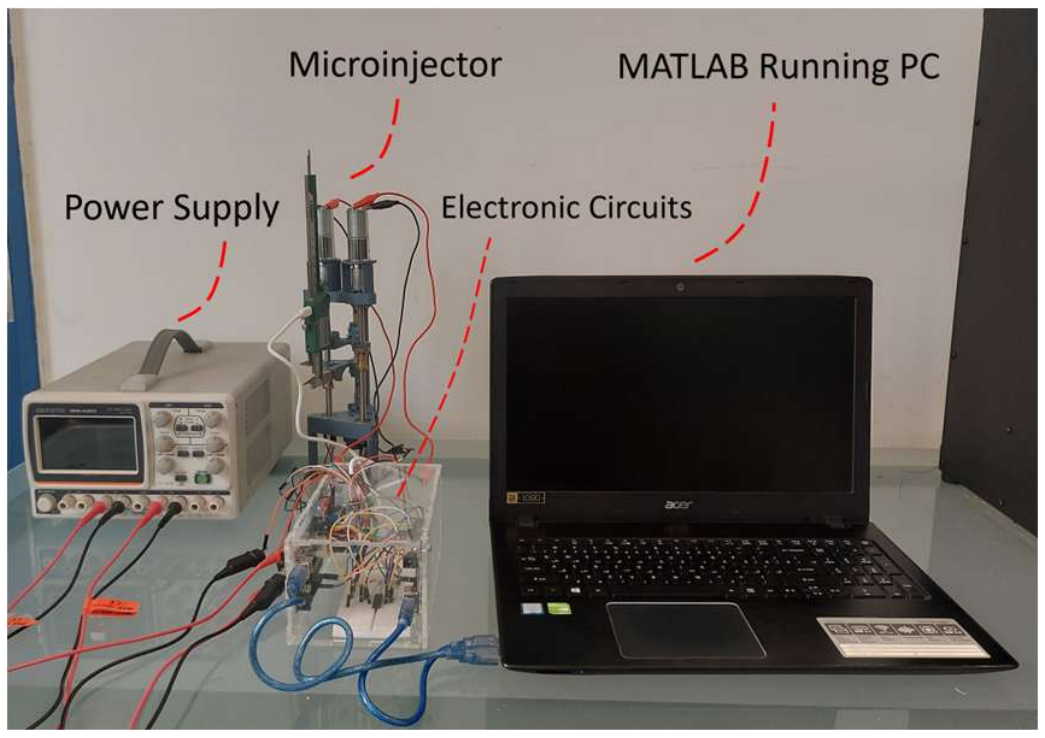

Figure 10. Final Injection Setup

- Enabling precise adjustment of injection parameters
- Providing real-time information such as the device's current status, parameters, remaining injection time, and alerts
- Allowing the medical team to easily communicate issues and requirements with the technical team

Eventually, the final setup is displayed in Figure 10.

## III. RESULTS & DISCUSSIONS

In this section, the results obtained from various experiments and evaluations of the injection device are analyzed. The optimal motor rotation speed range for the intended application, as determined by experiment, is between 0.5 to 1.5 rpm. As mentioned in (1), the difference in angular velocity between the two motors should be 0.6 rpm to achieve the desired injection rate.

To evaluate the system design, various tests under standard conditions are designed and implemented, and their output is recorded. As an example, one of the test results will be discussed. In this test, the angular velocities of the motors are selected as 0.8 and 1.4 rpm. The designed GUI based on these numbers creates a ramp input for each dynamic system to be followed. Figures 11.a and 11.b display system input, output and controller output of each DC Motor. As it appears, the implemented controller which its coefficients are tabulated in table 3, makes the systems to track their reference with a negligible steady state error. Figures 12.a and 12.b show the numerical derivative of encoder outputs of the first and second DC motors, respectively, for inputs corresponding to 1.4 and 0.8 rpm.

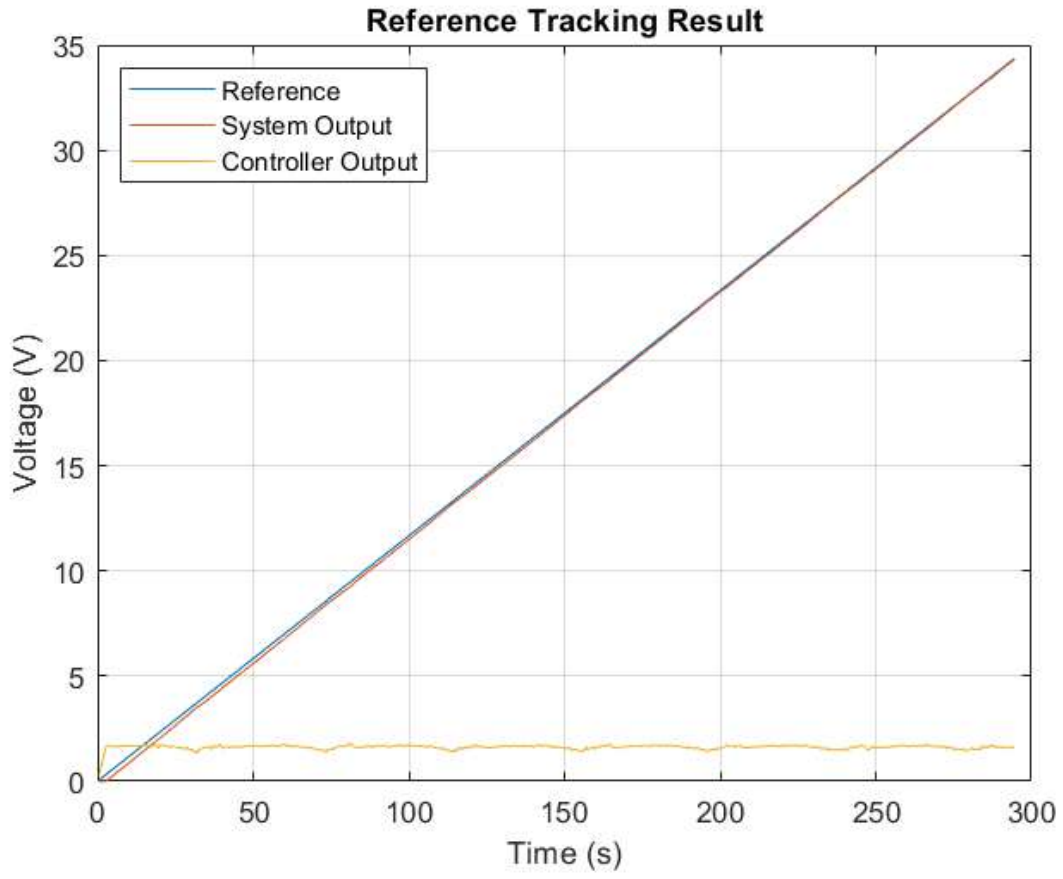

a. Body DC Motor

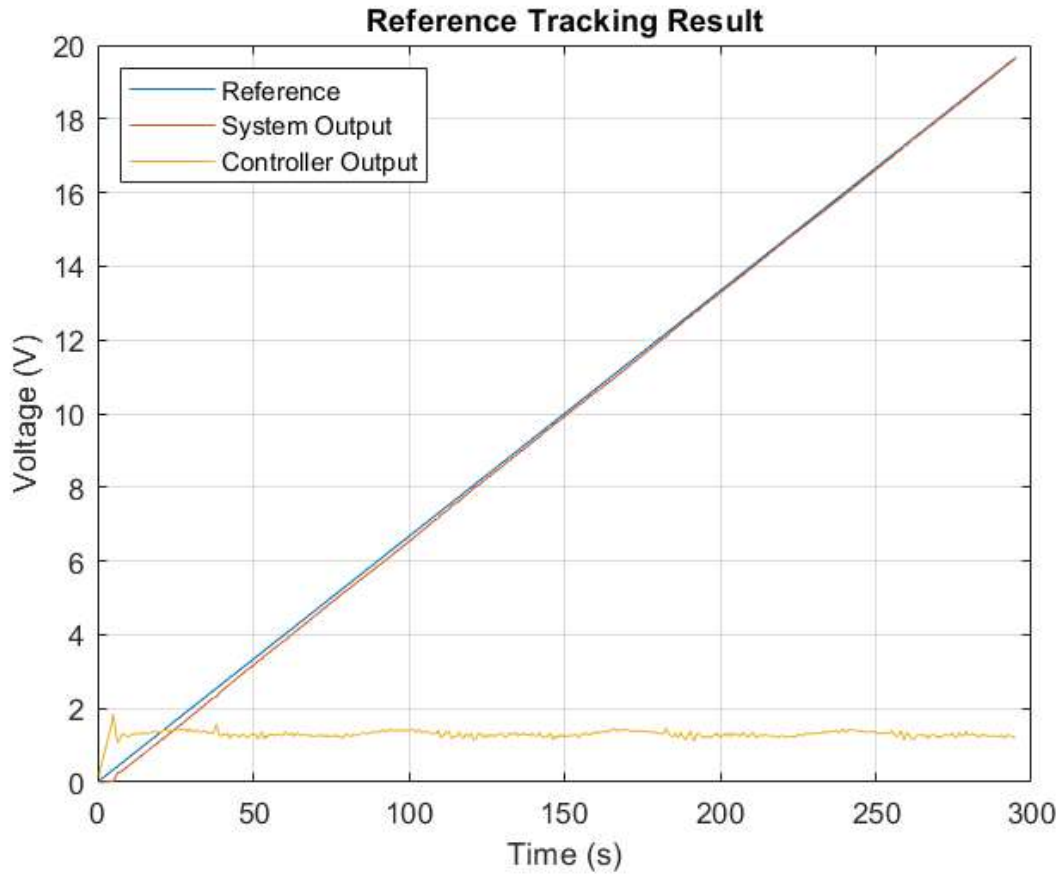

b. Plunger DC Motor

Figure 11. System Reference, Output, and Controller Output

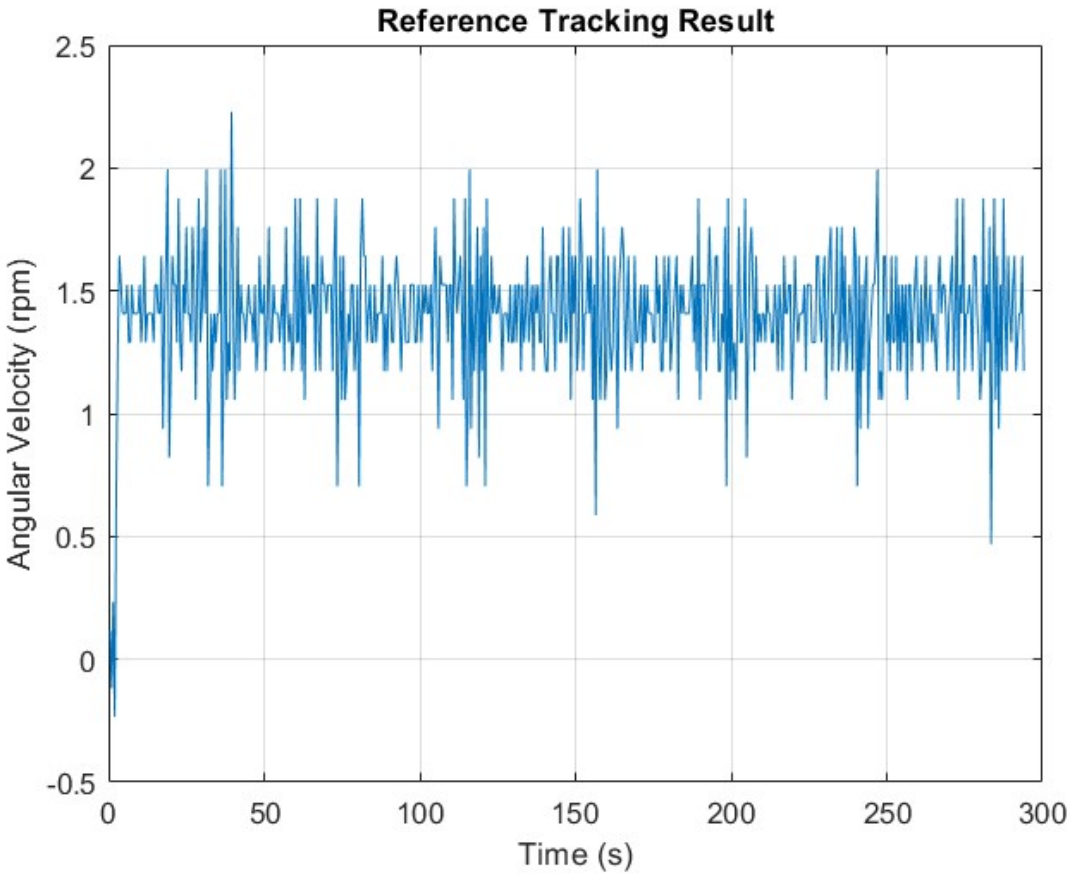

a. Body DC Motor

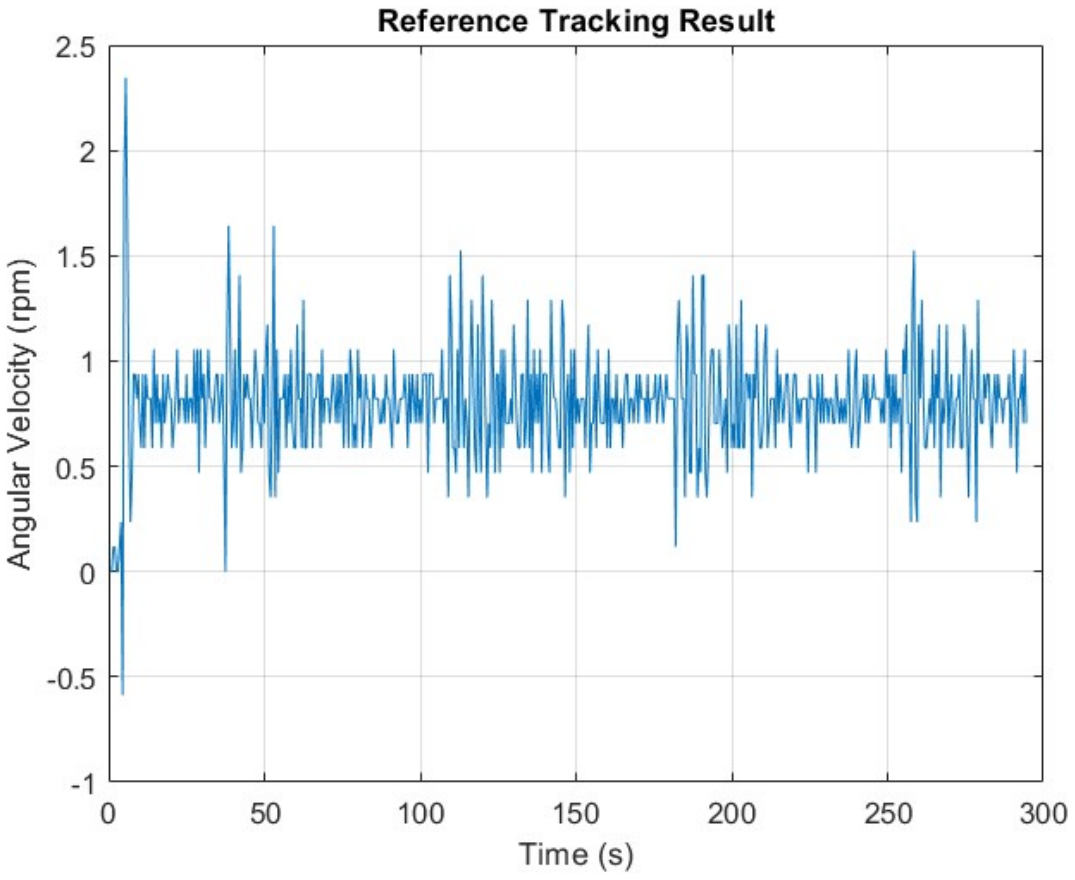

b. Plunger DC Motor

Figure 12. Angular Velocity of DC Motors Measured by Magnetic Encoders

By calculating the average and standard deviation of each motor's speed, a better understanding of the injection process's accuracy can be obtained. For the first motor, the average angular speed is 1.4 rpm with a standard deviation of 0.26 rpm. For the second motor, the average angular speed is 0.8 rpm with a standard deviation of 0.25 rpm. The average injection rate in this state is 0.99 microliters per minute, with a standard deviation of 0.56 microliters per minute. This result is purely obtained from the output of the magnetic encoders.

To evaluate the reliability of the encoders, the injection rate is once again calculated using the digital caliper output. The liquid injection rate over time, estimated by the outputs gathered from the digital caliper, is shown in Figure 13. The average injection rate is 0.99 microliters per minute, with a standard deviation of 0.18 microliters per minute.

The caliper's measurement results demonstrate that although the magnetic encoder's output is noisy, the mean value of its measurement is accurate and can be relied on. However, since the standard deviation of the injection rate obtained from the encoders is relatively high, the use of a more precise distance measurement sensor in the closed-loop control structure can enhance the results.

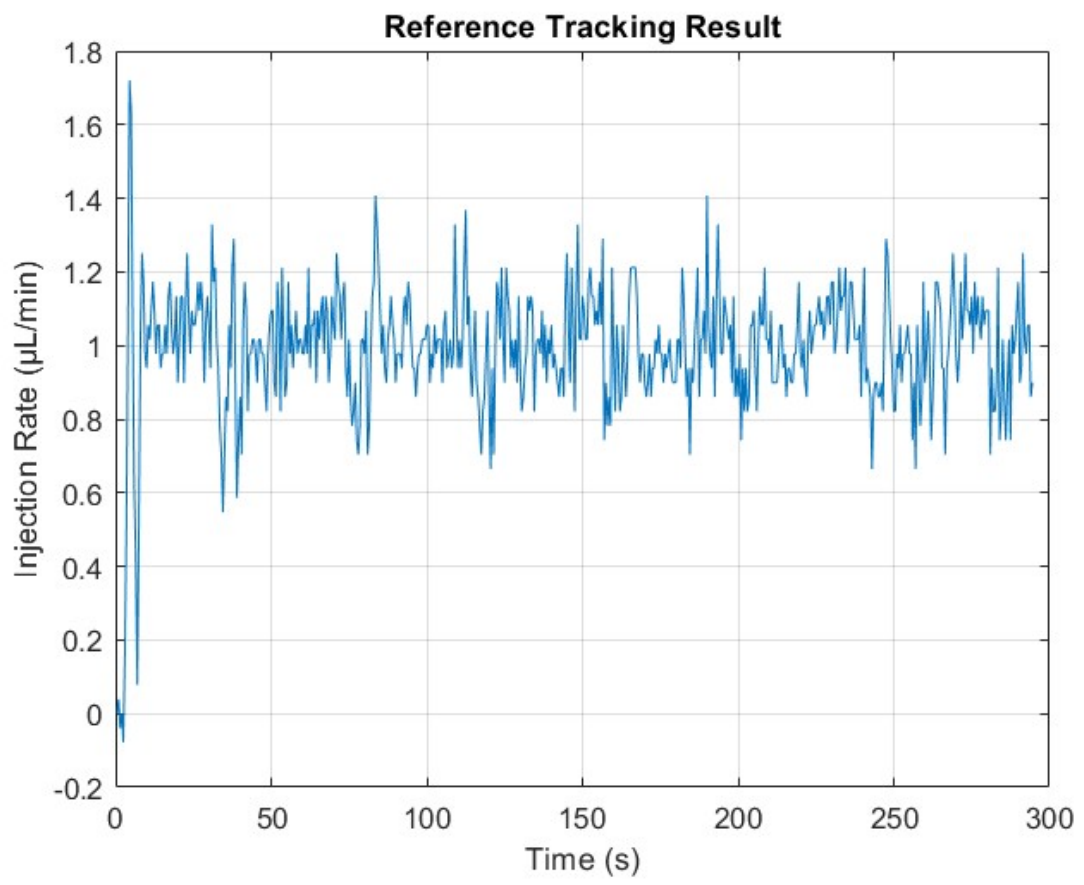

Figure 13. The Injection Rate of the Liquid Measured by The Digital Caliper

## IV. CONCLUSION

In this research, a microinjector has been designed and fabricated to conduct the columnar injection of stem cells into the patient's brain. This injection method has increased the success of therapeutic approaches by reducing damage to brain tissue. To achieve this, the injection rate must remain constant during the procedure. By utilizing two encoders and implementing a PID controller, operational errors and environmental noise have been significantly reduced, allowing the injection rate to be kept steady. The encoder data has indicated that the average difference between the actual injection rate and the desired rate is approximately one percent. Additionally, a digital caliper has been employed to validate the encoder data. The caliper data have shown an average injection rate similar to that of the encoder, but the standard deviation in the caliper data has been significantly lower. This indicates that the device has functioned appropriately and as expected, with the noise in the encoder data being attributed to sensor measurement.